\def\year{2022}\relax
\documentclass[letterpaper]{article} % DO NOT CHANGE THIS
\usepackage{aaai22}  % DO NOT CHANGE THIS
\usepackage{times}  % DO NOT CHANGE THIS
\usepackage{helvet}  % DO NOT CHANGE THIS
\usepackage{courier}  % DO NOT CHANGE THIS
\usepackage[hyphens]{url}  % DO NOT CHANGE THIS
\usepackage{graphicx} % DO NOT CHANGE THIS
\usepackage{natbib}  % DO NOT CHANGE THIS AND DO NOT ADD ANY OPTIONS TO IT
\usepackage{caption} % DO NOT CHANGE THIS AND DO NOT ADD ANY OPTIONS TO IT
\DeclareCaptionStyle{ruled}{labelfont=normalfont,labelsep=colon,strut=off} % DO NOT CHANGE THIS
\usepackage{algorithm}
\usepackage{algorithmic}
\usepackage{amssymb}
\usepackage{amsmath}
\usepackage{multirow}
\usepackage{newfloat}
\usepackage{listings}
\floatstyle{ruled}
\newfloat{listing}{tb}{lst}{}
\floatname{listing}{Listing}
\title{Dynamic Spatial Propagation Network for Depth Completion}
\author{
    Yuankai Lin\textsuperscript{\rm 1},
    Tao Cheng\textsuperscript{\rm 2}\thanks{Corresponding author},
    Qi Zhong\textsuperscript{\rm 3},
    Wending Zhou\textsuperscript{\rm 4} and
    Hua Yang\textsuperscript{\rm 1} 
}
\affiliations{
    \textsuperscript{\rm 1}School of Mechanical Science and Enginering, Huazhong University of Science and Technology, China\\
    \textsuperscript{\rm 2}College of Urban Transportation and Logistics, Shenzhen Technology University, China\\
    \textsuperscript{\rm 3}Division of Logistics and Transportation, Tsinghua University, China\\
    \textsuperscript{\rm 4}School of Science and Engineering, The Chinese University of Hong Kong (Shenzhen), China\\
    \{linyuankai,huayang\}@hust.edu.cn, chengtao@sztu.edu.cn, zhongq20@mails.tsinghua.edu.cn, 220019048@link.cuhk.edu.cn
}

\usepackage{bibentry}
\begin{document}

\maketitle

\begin{abstract}
Image-guided depth completion aims to generate dense depth maps with sparse depth measurements and corresponding RGB images. Currently, spatial propagation networks (SPNs) are the most popular affinity-based methods in depth completion, but they still suffer from the representation limitation of the fixed affinity and the over smoothing during iterations. 
%suffer from the same performance bottleneck in terms of linear propagation.  
Our solution is to estimate independent affinity matrices in each SPN iteration, but it is over-parameterized and heavy calculation.
This paper introduces an efficient model that learns the affinity among neighboring pixels with an attention-based, dynamic approach. Specifically, the Dynamic Spatial Propagation Network (DySPN) we proposed makes use of a non-linear propagation model (NLPM). It decouples the neighborhood into parts regarding to different distances and recursively generates independent attention maps to refine these parts into adaptive affinity matrices. 
%assigns different attention levels to neighboring pixels of different distances when refining each pixel in the depth map recursively.
Furthermore, we adopt a diffusion suppression (DS) operation so that the model converges at an early stage to prevent over-smoothing of dense depth. Finally, in order to decrease the computational cost required, we also introduce three variations that reduce the amount of neighbors and attentions needed while still retaining similar accuracy. In practice, our method requires less iteration to match the performance of other SPNs and yields better results overall. DySPN outperforms other state-of-the-art (SoTA) methods on KITTI Depth Completion (DC) evaluation by the time of submission and is able to yield SoTA performance in NYU Depth v2 dataset as well.
\end{abstract}

\begin{figure}[t]
\centering
\includegraphics[width=1.0\columnwidth]{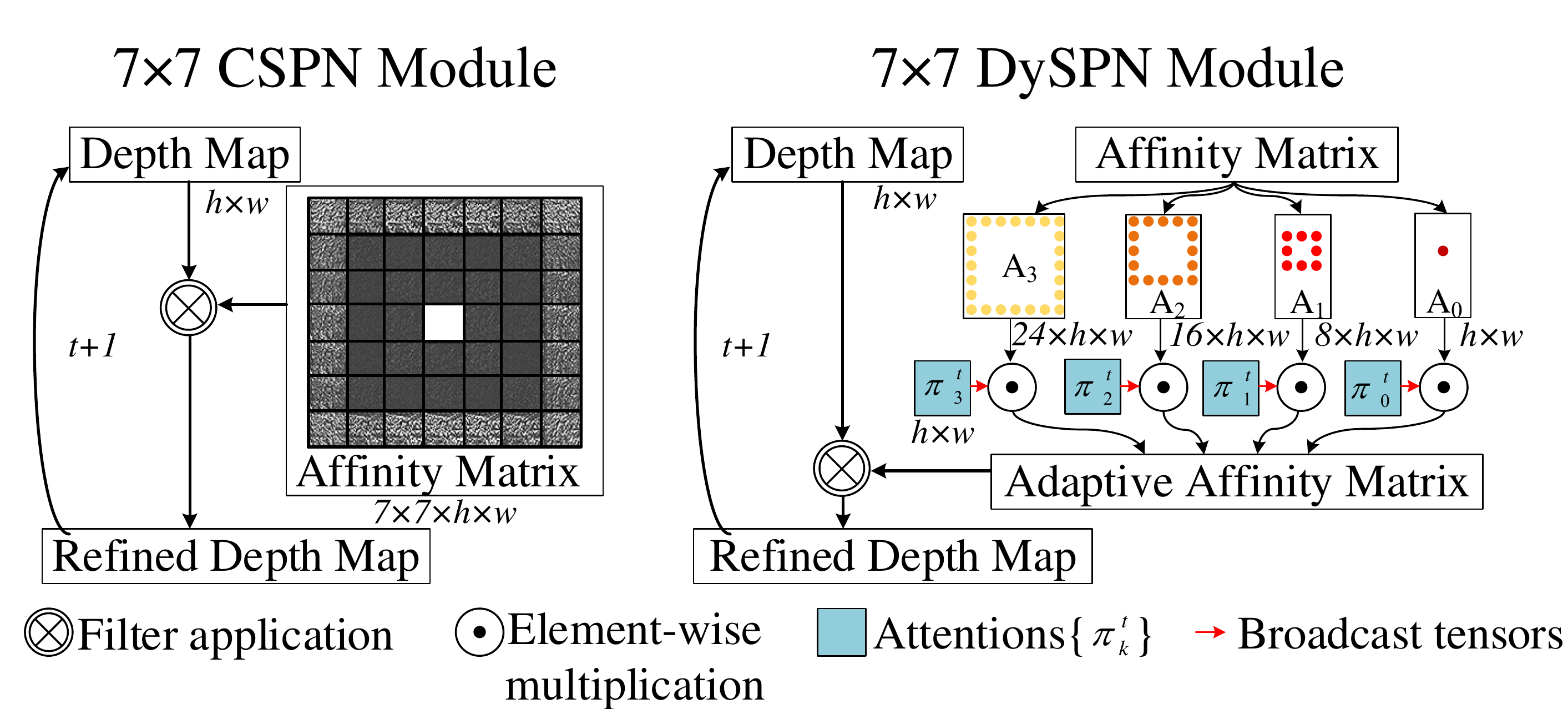} % Reduce the figure size so that it is slightly narrower than the column. Don't use precise values for figure width.This setup will avoid overfull boxes.
\caption{Comparison between 7×7 CSPN and 7×7 DySPN. Left: A fixed affinity matrix is assigned to CSPN, with its weights distributed in the outermost part of the kernel. Right: An adaptive affinity matrix adjusted by the attention mechanism is provided to our DySPN. }
\label{fig1}
\end{figure}

\section{Introduction}
Dense depth estimation is essential for 3D vision tasks, e.g., 3D object detection \cite{patchnet} and tracking \cite{Hu_2019_ICCV}, 6D object pose estimation \cite{DenseFusion}, simultaneously localization and mapping (SLAM) \cite{V-LOAM}, and structure-from-motion (SfM) \cite{8578578}, which are widely used in autonomous driving and robot perception. However, the output of depth sensors such as LiDAR and ToF cameras is sparse, resulting in a large number of empty data regions, which need further processing to fulfill or simply ignore these missing regions.

Since RGB images can reflect subtle changes of color and texture, many existing works \cite{8460184}  \cite{8757939} \cite{8765412} utilize synchronous RGB images as guidance to complete sparse depth values, which are known as image-guided depth completion. These methods directly estimate dense depths by RGB images and sparse depths and use encoder-decoder architecture \cite{7803544} to generate more details of depth maps. Therefore, compared with the early methods \cite{6126488} \cite{8374553} \cite{8575731} that only rely on sparse measurement, the results of depth completion are significantly improved. In order to generate better depth image boundaries, many learnable edge-preserving strategies \cite{8869936} \cite{guidenet} are adopted, which can be easily embedded into end-to-end deep convolutional neural networks. SPNs are based on the strategy of Anisotropic Diffusion \cite{Weickert96anisotropicdiffusion}, which are popular refinement techniques in depth estimation and depth completion. This strategy obtains the final dense depth maps by affinities of local \cite{CSPN++} or non-local \cite{NLSPN} neighbors and iteratively refines the depth prediction. However, most of the current SPNs are based on linear propagation models. The values of the affinity matrix would not change during the propagation process, which greatly limits the representation capability of SPNs. Taking CSPN \cite{CSPN} as an example, as illustrated in Fig. \ref{fig1}, when the kernel size is 7, most of the affinity weights are distributed in the distance, which means that the affinity matrix cannot clearly represent the relationship between pixels and neighbors.  

To solve the problems mentioned above, inspired by Dynamic filters \cite{DFN}, we proposed a new non-linear propagation design, named Dynamic Spatial Propagation Network (DySPN), which can learn an adaptive affinity matrix. To implement this design, different attention levels are assigned to neighbors of different distances when refining each pixel in the depth map recursively. More specifically, at the beginning of the propagation, far neighbors provide long-range information to fill the hole of the initial depth map and smooth it quickly. As the depth map becomes denser, near neighbors are paid more attention to edge-preserving. Our attention mechanism adjusts the affinity matrix according to each propagation stage, making the process more accurate and efficient. Besides, we adopt a diffusion suppression operation, which estimates the affinities of refining depths in every step to preserve good parts of the depths and avoid over-smoothing. The affinity of refining depth is easy to participate in the propagation process with nearly no additional computational cost. Although the computational complexity of our DySPN is similar to CSPN, we try to use a larger receptive field and fewer neighbors to achieve a better result. Thus, we introduce three variations that reduce the amount of distant pixels needed dramatically while still retaining similar accuracy. Compared with other SPNs, the iteration steps of our proposed method is only half, which is 6 in this paper. 

In summary, our contribution lies in three aspects:
\begin{itemize}
\item 
%We propose a non-linear propagation model named DySPN. The adaptive affinity matrices can better reflect the relationship between pixels and their neighbors, and adjust dynamically at pixel level during the spatial propagation processing. 
We propose a non-linear propagation model (NLPM) named DySPN. This model adjusts affinity weights dynamically which is a new prospect to improve SPN. We further achieve it by neighborhood decoupling and spatial-sequential attention.   
\item A diffusion suppression (DS) operation is proposed, which can adaptive terminate the propagation and reduce over-smoothing.
\item Three implementations of the DySPN model are proposed, which can effectively reduce computational complexities and improve performance.
\end{itemize}

\begin{figure*}[t]
\centering
\includegraphics[width=0.9\textwidth]{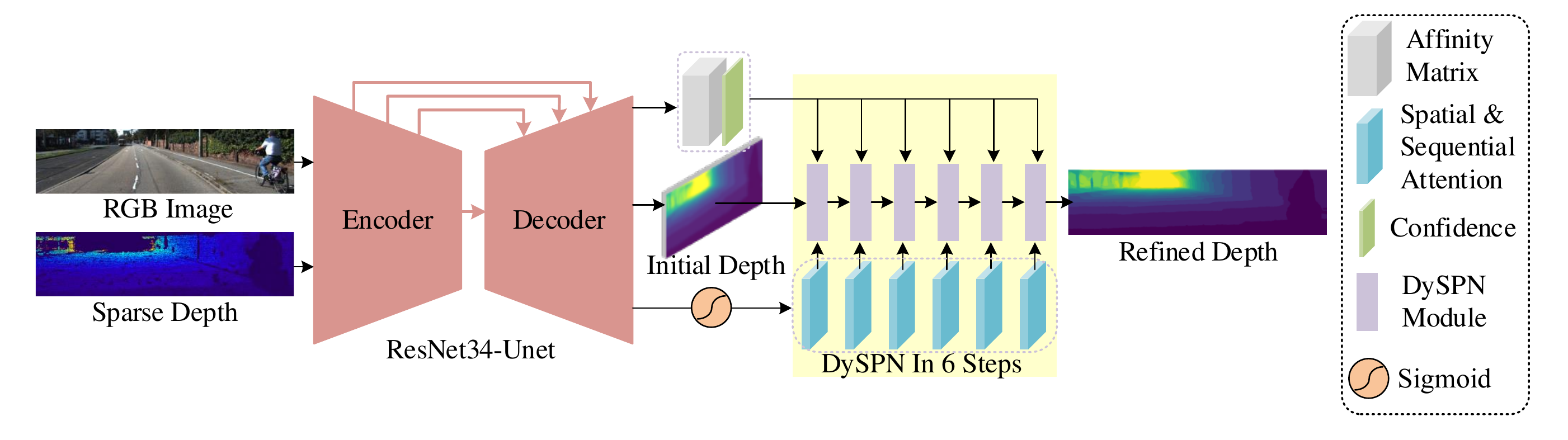} % Reduce the figure size so that it is slightly narrower than the column.
\caption{The overall framework of our networks with DySPN. We use a simple ResNet34-Unet to generate an initial depth map, an affinity matrix, a serial of spatial and sequential attention maps. The attention maps are split into six parts and applied to DySPN module in a 6-steps propagation process.}
\label{fig2}
\end{figure*}

\section{Related Work}
%\textbf{Image-guided depth completion.}
\paragraph{Image-guided depth completion} Early depth completion methods  \cite{6126488} \cite{6335696} \cite{6865733} only rely on sparse measurements to reconstruct or fill holes. image-guided depth Completion methods \cite{4270236} \cite{6126423} \cite{8460184} usually achieve better results because the additional RGB images contain a large amount of surface, edge, or semantic information. 

Later works conduct the guidance or fusion strategies of these RGB image information. For example, DeepLiDAR \cite{8953782} is a network composed of a color pathway and a surface normal pathway. ACMNet \cite{9440471} adopts the graph propagation to capture the observed contextual information in the encoder stage. Inspired by the guided image filtering, GuideNet \cite{guidenet} proposes an image guidance convolution module for multi-modal feature fusion. FCFR-NET \cite{liu2021fcfr} designs an energy-based fusion operation to fuse the features of color and depth information. PENet \cite{hu2020PENet} uses addition operation to merge color and depth features of different modalities at multiple stages. Besides, KBNet \cite{wong2021unsupervised} transfers spatial positional encoding across different branches of the encoder by a calibrated backprojection module.

These methods usually use two independent branches to extract the features of RGB images and sparse depth maps and fuse them at different scales.
%\textbf{Spatial propagation networks.}
\paragraph{Spatial propagation networks} SPNs are depth refinement technologies that differ from the aforementioned two-branch architectures. They are usually connected by a simple Unet \cite{RFB15a} network. Among them, SPN \cite{SPN} is the first proposed method to use a deep CNN to learn an affinity matrix that effectively models any pairwise similarity matrix. SPN builds a row/column linear propagation model whose spatially varying transformation matrix constitutes the affinity matrix. It can be easily applied to many high-level vision tasks, including depth completion and semantic segmentation. Later, CSPN \cite{CSPN} further improves the linear propagation model and adopts a recursive convolution operation to be more efficient. CSPN++ \cite{CSPN++} merges the outputs of three independent CSPN modules so that its propagation learns adaptive convolutional kernel sizes and the number of iterations. NLSPN \cite{NLSPN} predicts the relevant non-local neighbors to avoid the mixed-depth problem caused by irrelevant local neighbors. DSPN \cite{9191138} utilizes their deformable SPN that adaptively adjusts receptive fields at each pixel. ODE-CNN\cite{9197123} adopts an inverse gnomonic projection CSPN and a deformable CSPN, which better recovers the structural details of Omnidirectional depth maps. PENet \cite{hu2020PENet} implement a dilated CSPN++ to enlarge the propagation neighborhoods. 

However, these methods are constrained by the bottleneck of the linear propagation model, as the affinity matrix is fixed during propagation. Fixed affinity weights make it difficult for propagation to take into account both long-range dependencies and local information, resulting in more SPN iterations. 
%\textbf{Dynamic Filters.}
\paragraph{Dynamic filters} Dynamic filters \cite{DFN} \cite{8954063} could dynamically modify or predict filter weights based on the input features, and has shown a strong connection to the attention mechanism \cite{Xu2016AskAA}. CondConv \cite{NEURIPS2019_f2201f51} parameterizes several convolutional kernels as a linear combination to mix the experts voted. WeightNet \cite{WeightNet} directly generates the weights by a grouped fully-connected layer which is different from common practice that applies operations on vectors in feature space. DyNet \cite{zhang2020dynet} presents a dynamic convolution kernel generation method whose coefficient prediction module is trainable and designed to predict the coefficients of fixed convolution kernels. DynamicConv \cite{9157588} utilizes an Attention mechanism to aggregate multiple convolution kernels to increase the model capability. Furthermore, DDF \cite{zhou_ddf_cvpr_2021} simultaneously applies spatial and channel-wise attention to the dynamic filter of each pixel.

Inspired by the dynamic filters described above, we propose a DySPN with a non-linear propagation model, which applies spatial and sequential attention to generate a series of adaptive affinity matrices.

\section{Our Approach}
In this paper, we propose DySPN for image-guided depth completion, including a naive non-linear propagation model, a diffusion suppression operation, and three variants that greatly reduce the complexities. The pipeline of our approach is demonstrated in Fig. \ref{fig2}.
\subsection{Naive Non-linear Propagation Model}
We design a non-linear propagation model (NLPM) that satisfies the desired requirements of SPN \cite{SPN}. Assuming $h^{0}\in \mathbb{R}^{m\times n}$ is the initial depth map and $h^{N}\in \mathbb{R}^{m\times n}$ is the refined depth map after N iterations. $\{h^{0}$, $h^{N}\}$is reshaped as a column-first one-dimensional vector $V^{t}\in \mathbb{R}^{mn}$, where $t\in\{0,1,2\cdots N\}$. The NLPM could be simplely described as follows,
\begin{equation} 
\begin{split}
V^{t+1}&=GV^{t}\\
&=\begin{bmatrix}
1-\lambda_{0,0} & w_{0,0}(1,0) & \cdots  &w_{0,0}(m,n) \\ 
w_{1,0}(0,0) & 1-\lambda_{1,0} & \cdots & w_{1,0}(m,n)\\ 
\vdots & \vdots & \ddots  & \vdots\\ 
w_{m,n}(0,0) & w_{m,n}(1,0) & \cdots & 1-\lambda_{m,n}
\end{bmatrix}V^{t}
\end{split}
\end{equation}
where $G\in \mathbb{R}^{mn\times mn}$ is a fixed transformation matrix during the propagation, $w_{i,j}(a,b)$ describes the affinity weight of pixel$(i, j)$ with its neighbors$(a, b)$ and $\lambda _{i,j}=\sum _{a\neq i,b\neq j}w_{i,j}(a,b)$ means that the sum of each row of G is 1. 

Ideally, our method adjusts the value of G by applying global attention. Therefore, the NLPM corresponds to the diffusion process with a partial differential equation (PDE) in Eq.(2).
\begin{equation}
\begin{split}
V^{t+1}&=G^{t}V^{t}=(I-D^{t}+K^{t}\cdot A)V^{t} \\
V^{t+1}-V^{t}&=-(D^{t}-K^{t}\cdot A)V^{t}\\ 
\partial _{t}V^{t+1}&=-L^{t}V^{t}
\end{split}
\end{equation}
Where $K^{t}\in \mathbb{R}^{mn\times mn}$ is the global attention matrix, $D$ is the diagonal matrix that contains all the $\lambda _{i,j}$, and $A$ is the affinity matrix containing all the  $w_{i,j}(a,b)$. 

$L^{t}$ is the Laplace matrix that varies with the number of iterations $t$, indicating that our model is non-linear. The transformation matrix $G^{t}$ and  the global attention matrix $K^{t}$ could be concretely written as follows, 
\begin{equation} 
\begin{split}
G^{t}=\begin{bmatrix}
1-\tilde{\lambda}_{0,0}^{t} & \tilde{w}_{0,0}^{t}(1,0) & \cdots  &\tilde{w}_{0,0}^{t}(m,n) \\ 
\tilde{w}_{1,0}^{t}(0,0) & 1-\tilde{\lambda}_{1,0}^{t} & \cdots & \tilde{w}_{1,0}^{t}(m,n)\\ 
\vdots & \vdots & \ddots  & \vdots\\ 
\tilde{w}_{m,n}^{t}(0,0) & \tilde{w}_{m,n}^{t}(1,0) & \cdots & 1-\tilde{\lambda}_{m,n}^{t}
\end{bmatrix}
\end{split}
\end{equation}
\begin{equation} 
\begin{split}
K^{t}=\begin{bmatrix}
0 & \pi_{0,0}^{t}(1,0) & \cdots  &\pi_{0,0}^{t}(m,n) \\ 
\pi_{1,0}^{t}(0,0) & 0 & \cdots & \pi_{1,0}^{t}(m,n)\\ 
\vdots & \vdots & \ddots  & \vdots\\ 
\pi_{m,n}^{t}(0,0) & \pi_{m,n}^{t}(1,0) & \cdots & 0
\end{bmatrix}
\end{split}
\end{equation}
where $\tilde{w}_{i,j}^{t}(a,b)=\pi_{i,j}^{t}(a,b)w_{i,j}(a,b)$ is the weight of adaptive affinity, and $\tilde{\lambda }_{i,j}^{t}=\sum _{a\neq i,b\neq j}\tilde{w}_{i,j}^{t}(a,b)$.
\subsection{Diffusion Suppression Operation}
The diffusion process would not stop until the end of N iterations, but this is inappropriate for all pixels. If N is too high, the output will be over-smoothing, which is negative for depth completion. To solve this problem, we adopt a diffusion suppression operation that estimates the affinity of refined depth maps in every iteration so that the model converges at an early stage. 

We use a diagonal matrix $\bar{D}^{t}$ to describe the affinity of refined depth maps between $t$ and $t+1$ step in Eq.(5). 
\begin{equation} 
\begin{split}
\bar{D}^{t}=I-D^{t}=\begin{bmatrix}
1-\tilde{\lambda}_{0,0}^{t} & 0 & \cdots  & 0 \\ 
0 & 1-\tilde{\lambda}_{1,0}^{t} & \cdots & 0\\ 
\vdots & \vdots & \ddots  & \vdots\\ 
0 & 0 & \cdots & 1-\tilde{\lambda}_{m,n}^{t}
\end{bmatrix}
\end{split}
\end{equation}
When the diffusion is completely convergent, the attention matrix $K^{t}$ is a zero matrix, deducing $\tilde{\lambda }_{i,j}^{t}=0$. The diffusion process suppress as follows.
\begin{equation} 
\begin{split}
V^{t+1}=G^{t}V^{t}=\bar{D}^{t}V^{t}=IV^{t}=V^{t}
\end{split}
\end{equation}
\subsection{Neighborhood decoupling of DySPN}
Our basic idea of $N$-step NLPM uses global attention with a large complexity of $O(m^{2}n^{2}N)$. When we individually assign attention to each sampled neighbor, the complexity drops to $O(k^{2}N)$. Meanwhile, the base network needs to estimate N times more feature maps than the linear propagation model, accompanied by a large number of additional calculations. 

\begin{figure}[t]
\centering
\includegraphics[width=1.0\columnwidth]{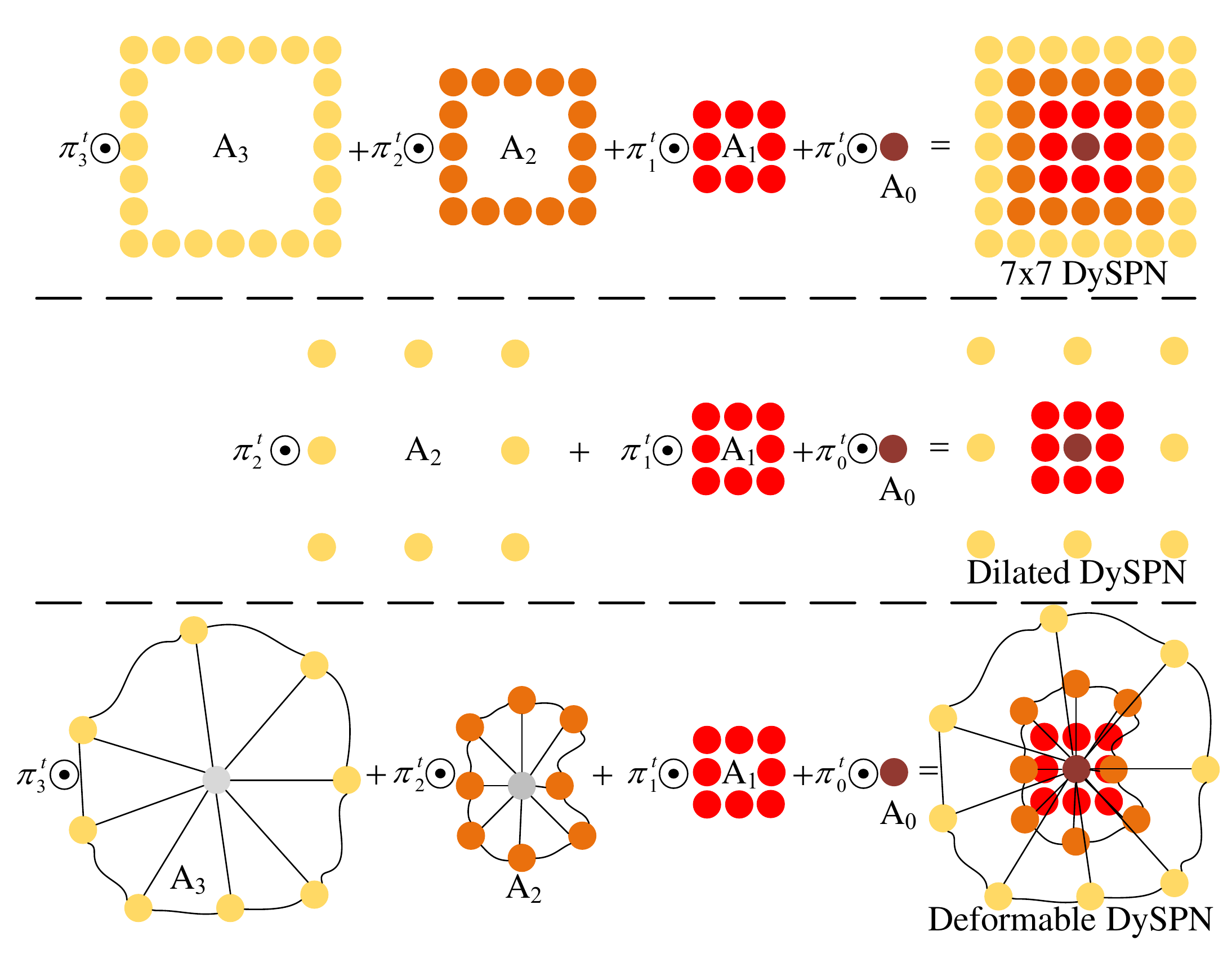} % Reduce the figure size so that it is slightly narrower than the column. Don't use precise values for figure width.This setup will avoid overfull boxes.
\caption{Neighborhood decoupling of three variants. Top: 7×7 DySPN decomposes the neighborhood into four parts based on distance. Middle: Dilated DySPN aggregates two different 3×3 neighborhoods, one of which has a dilation of 3. Bottom: Deformable DySPN uses two deformable 3×3 neighborhoods to obtain different receptive fields. }
\label{fig3}
\end{figure}

By analyzing the 7×7 CSPN affinity matrix, we find that the affinity weights of the neighbors at the same distance are correlated. Therefore, we came up with three variants for our DySPN, which decouple the neighborhood into parts regarding different distances. The complexity of our decoupled attention mechanism is only $O(kN)$, which is irrelevant to image size$(m, n)$. Our method is easy to be implemented and lightly embedded in the propagation. As illustrated in Fig. \ref{fig3}, we reduce the amount of neighbors needed while achieving better accuracy in our experiment. Generally, one step propagation of our DySPN could be written as,
\begin{equation} 
\begin{split}
%h_{i,j}^{t+1}=&\frac{\pi_{i,j,3}^{t}}{{S}'_{i,j}}\sum_{(a,b)\in N_{i,j,3}^{t}}w_{i,j}(a,b)h_{a,b}^{t}\\
%             +&\frac{\pi_{i,j,2}^{t}}{{S}'_{i,j}}\sum_{(a,b)\in N_{i,j,2}^{t}}w_{i,j}(a,b)h_{a,b}^{t}\\
%             +&\frac{\pi_{i,j,1}^{t}}{{S}'_{i,j}}\sum_{(a,b)\in N_{i,j,1}^{t}}w_{i,j}(a,b)h_{a,b}^{t}\\
%             +&\frac{\pi_{i,j,0}^{t}}{{S}'_{i,j}}h_{i,j}^{t}+(1-\frac{S_{i,j}}{{S}'_{i,j}})h_{i,j}^{0}
h_{i,j}^{t+1}=&\sum_{k\in \mathbb{Z}_{+}}\sum_{(a,b)\in N_{i,j,k}^{t}}\frac{\pi_{i,j,k}^{t}}{{S}'_{i,j}}w_{i,j}(a,b)h_{a,b}^{t}\\
             +&\frac{\pi_{i,j,0}^{t}}{{S}'_{i,j}}h_{i,j}^{t}+(1-\frac{S_{i,j}}{{S}'_{i,j}})h_{i,j}^{0}
\end{split}
\label{eq7}
\end{equation}
where,
\begin{equation} 
\begin{split}
S_{i,j}&=\pi_{i,j,0}^{t}+\sum_{k\in \mathbb{Z}_{+}}\sum_{(a,b)\in N_{i,j,k}^{t}}\pi_{i,j,k}^{t}w_{i,j}(a,b),\\
{S}'_{i,j}&=\pi_{i,j,0}^{t}+\sum_{k\in \mathbb{Z}_{+}}\sum_{(a,b)\in N_{i,j,k}^{t}}\pi_{i,j,k}^{t}|w_{i,j}(a,b)|\nonumber.
\end{split}
\end{equation}

The affinity matrix $A$ is usually extremely sparse because only a limited number of neighbors are sampled. Therefore, we use $N_{i,j,k}^{t}$ to clearly represent its decoupled neighborhood. The basic form of $N_{i,j,k}^{t}$ is defined as:
\begin{equation} 
\begin{split}
N_{i,j,k}^{t}=\{(i+p,j+q)|p=\pm k\ or\ q=\pm k,\\
(-k,-k)\leqslant (p,q)\leqslant(k,k),(p,q)\in \mathbb{R}\}\nonumber.
\end{split}
\end{equation}
where $h_{i,j}^{t}$ is the pixel$(i,j)$ value of $h^{t}$, and $N_{i,j,k}^{t}$ is the set of neighbors at pixel distance $k\in \mathbb{Z}_{+}$. $w_{i,j}(a,b)$ is the affinity matrix weight between pixel $(i,j)$ and its neighbour $(a,b)$. Notice that the weight of affinity matrix is normalized by $S_{i,j}$. The spatial and sequential attentions $\pi_{i,j,k}^{t}$ are activated by a sigmoid function and incorporated into the propagation.Specifically, when the kernel is of size 7, a 7×7 DySPN is shown as the top of Fig. \ref{fig3}. 

Dilated DySPN uses fewer neighbors to get a similar receptive field, as shown below:
\begin{equation} 
\begin{split}
N_{i,j,k}^{t}=\{(i+p,j+q)|p\in \{-2k+1,0,2k-1\},\\
q\in \{-2k+1,0,2k-1\},(p,q)\neq (0,0)\}\nonumber \\
\end{split}
\end{equation}

At the bottom of Fig. \ref{fig3}, Deformable DySPN has a trade-off between the number of neighbors and the size of receptive fields. Its neighborhood in pixel$(i,j)$ is defined as follows,
\begin{equation} 
\begin{split}
N_{i,j,1}^{t}=&\{(i+p,j+q)|p\in\{1,0,-1\},q\in\{1,0,-1\},
\\&(p,q)\neq (0,0)\}\\
N_{i,j,k}^{t}=&\{(i+p,j+q)|(p,q)\in f_{\phi ,k}(i,j),(p,q)\in \mathbb{R},\\&k\in\{2,3\}\}\nonumber 
\end{split}
\end{equation}
where $f_{\phi }$ estimates the off-set of neighbors for pixel$(i,j)$. 
	
\begin{table}[t]
\centering
\begin{tabular}{l|lllll}
\hline
\multirow{2}{*}{Method}& RMSE & REL & $\delta _{1.25}$ & $\delta _{1.25^{2}}$ & $\delta _{1.25^{3}}$\\ 
& \ (m) &  &  &  & \\ 
\hline
%S2D\_18 & 0.230 & 0.044  & 97.1 & 99.4 &99.8 \\ 
CSPN&  0.117 & 0.016 & 99.2 & \textbf{99.9} & \textbf{100.0} \\ 
CSPN++ & 0.116  & - & - & - & -\\ 
DeepLiDAR & 0.115 & 0.022 &99.3 &\textbf{99.9} &\textbf{100.0} \\ 
%DepthNormal &0.112& 0.018& 99.5& \textbf{99.9}& \textbf{100.0} \\ 
FCFRNet &0.106& 0.015& 99.5&\textbf{99.9}& \textbf{100.0} \\ 
ACMNet & 0.105& 0.015& 99.4& \textbf{99.9}& \textbf{100.0}\\ 
GuideNet & 0.101 &0.015 &99.5 &\textbf{99.9} &\textbf{100.0} \\
TWISE&0.097 & 0.013 &\textbf{99.6} &\textbf{99.9} &\textbf{100.0} \\
NLSPN&0.092& \textbf{0.012}& \textbf{99.6}& \textbf{99.9}& \textbf{100.0}\\
\hline
7×7 DySPN &\textbf{0.090}& \textbf{0.012}& \textbf{99.6}& \textbf{99.9}& \textbf{100.0}\\
Dilated -&\textbf{0.091}& \textbf{0.012}& \textbf{99.6}& \textbf{99.9}& \textbf{100.0}\\
Deformable -&\textbf{0.090}& \textbf{0.012}& \textbf{99.6}& \textbf{99.9}& \textbf{100.0}
\end{tabular}
\caption{Quantitative evaluation on NYU Depth v2 dataset. The CSPN, CSPN++, DeepLiDAR, FCFRNet, ACMNet, GuideNet, TWISE, NLSPN mean \cite{8460184}  \cite{CSPN} \cite{CSPN++} \cite{8953782} \cite{liu2021fcfr} \cite{9440471}  \cite{guidenet} \cite{TWISE} \cite{NLSPN}, respectively. }
\label{table2}
\end{table}
\begin{figure*}[t]
\centering
\includegraphics[width=2.1\columnwidth]{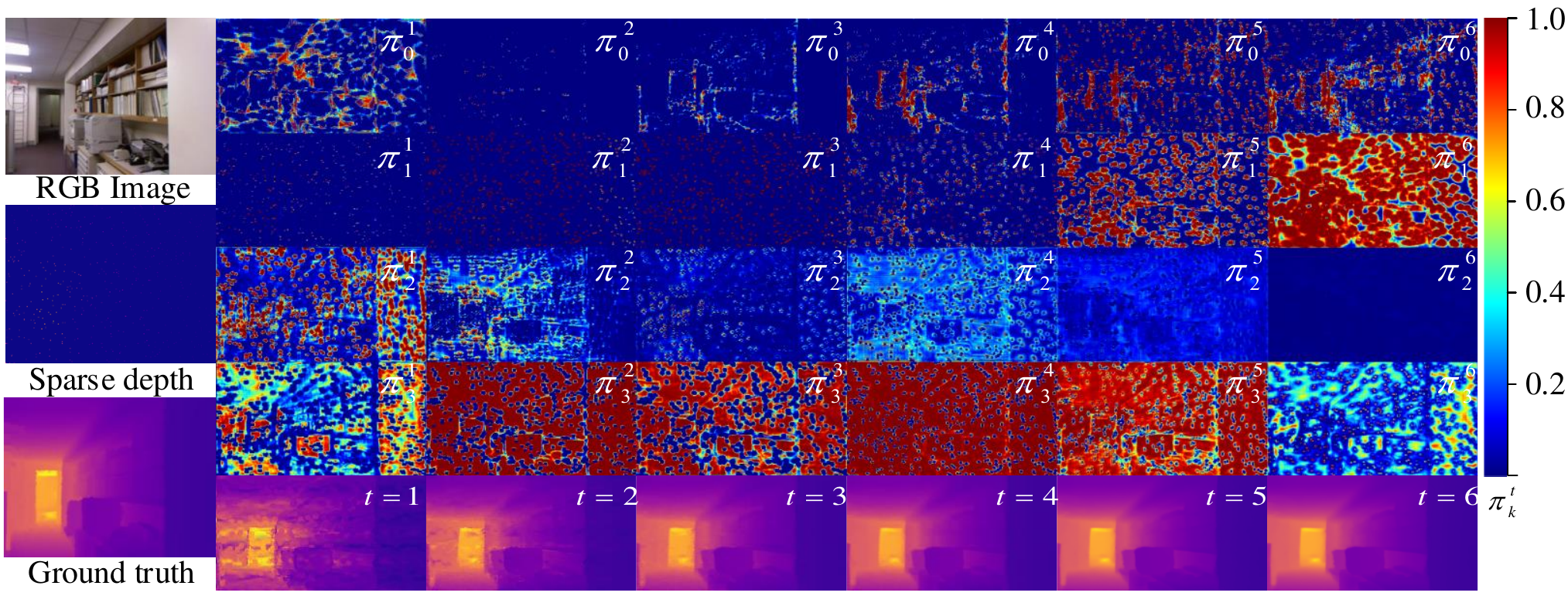} % Reduce the figure size so that it is slightly narrower than the column. Don't use precise values for figure width.This setup will avoid overfull boxes.
\caption{Visualization of attention maps $\pi_{k}^{t}$ activated by sigmoid functions in $7\times 7$ DySPN Modules. }
\label{fig4}
\end{figure*}
\subsection{Network Architecture}
An Unet-like \cite{RFB15a} encoder-decoder network with the backbone of ResNet-34 \cite{7780459} is proposed as a base network for fair comparison with other SPNs. The head of the base network is modified to fit our DySPN, so that it outputs attention maps $\pi_{k}^{t}$, an initial depth map $h^{0}$, an affinity matrix $A$, and a confidence prediction of the sparse depth measurement in parallel. The confidence prediction is common in other works such as \cite{CSPN++} \cite{9197123} \cite{hu2020PENet}. 
\subsection{Loss Function}
$L_{1}$ and $L_{2}$ loss are used at the same time to improve the performance of DySPN prediction results. Our loss is defined as:
\begin{equation} 
\begin{split}
Loss(h^{gt},h^{N})&=\alpha L_{1}(h^{gt},h^{N})+\beta L_{2}(h^{gt},h^{N})\\
L_{\rho }&=\frac{1}{S}\sum_{i}^{m}\sum_{j}^{n}|(h_{i,j}^{N}-h_{i,j}^{gt})\cdot \mathbb{I}(h_{i,j}^{gt}> 0)|^{\rho }\\
S&=\sum_{i}^{m}\sum_{j}^{n}\mathbb{I}(h_{i,j}^{gt}> 0)
\end{split}
\end{equation}
where $h^{N}$ is the N-steps refined depth map and $h^{gt}$ is the ground truth; $\mathbb{I}(h_{i,j}^{gt}> 0)$ is an indicator for the validity of $h^{gt}$ at pixel$(i,j)$; $\alpha=\beta=1$ and $\rho \in \{1,2\}$.

\section{Experiment}
%In this section, we first introduce the datasets we used, evaluation metrics, and implementation details. Next, we evaluate the performance of our method against other SoTA approaches on public available datasets, including the NYU Depth v2 dataset and KITTI DC dataset. Finally, Ablation studies are presented to verify the effectiveness of our method.
In this section, we follow the same experiment settings with other SPN-based methods (i.e. perform experiments on KITTI DC and NYU Depth v2 dataset) to have a fair comparison. Ablation studies are presented to verify the effectiveness of our method.

\subsection{Implementation Details}
We implement our method on the PyTorch framework and train it with 4 NVIDIA RTX 2080 Ti GPUs. All of our experiments are trained by an ADAM optimizer with $\beta_{1}$=0.9, $\beta_{2}$=0.999. The learning rate starts at 0.001 for the first 30 epochs and reduces to 0.002 with another 10 epochs of training. Furthermore, we use a stochastic depth \cite{Stochastic-Depth} strategy instead of the weight decay to prevent over-fitting during training for better performance. For the NYU Depth V2 dataset, the batch size is set to 24 and 500 depth pixels were randomly sampled from the ground truth depth. For the KITTI DC dataset, the batch size is 8. Besides, data augmentation techniques are used,  including horizontal random flip and color jitter.
\begin{figure*}[t]
\centering
\includegraphics[width=2.1\columnwidth]{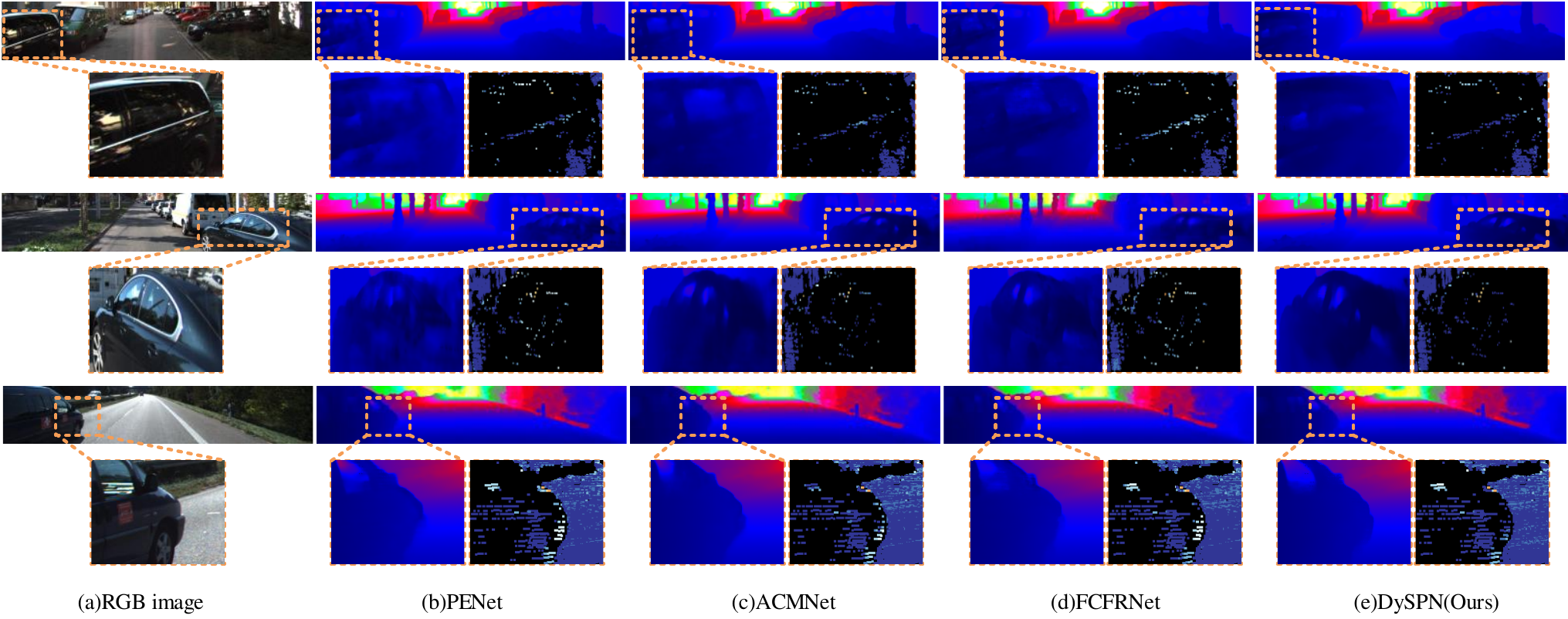} % Reduce the figure size so that it is slightly narrower than the column. Don't use precise values for figure width.This setup will avoid overfull boxes.
\caption{Qualitative comparisons results with other methods on KITTI DC evaluation. (b) PENet \cite{hu2020PENet}, (c) ACMNet \cite{9440471}, (d) FCFRNet \cite{liu2021fcfr}. }
\label{fig5}
\end{figure*}
\begin{table}[t]
\centering
\begin{tabular}{l|lllll}
\hline
\multirow{2}{*}{Method}& RMSE & MAE & iRMSE & iMAE\\ 
& (mm) & (mm)  & (1/km)  & (1/km)  \\ 
\hline
CSPN&  1019.64 & 279.46 & 2.93 & 1.15 \\ 
TWISE& 840.20 & 195.58 & 2.08 & \textbf{0.82}\\
%S2D&814.73 & 249.95 & 2.80 & 1.21\\
%DepthNormal &777.05 & 235.17 & 2.42 & 1.13 \\ 
DSPN&766.74 & 220.36 & 2.47 & 1.03\\
DeepLiDAR &758.38 & 226.50 & 2.56 & 1.15  \\ 
ACMNet &744.91 & 206.09 & 2.08 & 0.90\\ 
CSPN++ & 743.69 & 209.28 & 2.07 & 0.90\\ 
NLSPN&741.68 & 199.59 & 1.99 & 0.84\\
GuideNet &736.24 & 218.83 & 2.25 & 0.99 \\
FCFRNet &735.81 & 217.15 & 2.20 & 0.98\\ 
PENet &730.08 & 210.55 & 2.17 & 0.94\\ 
\hline
\multicolumn{1}{l|}{\begin{tabular}[l]{@{}l@{}}Deformable\\ DySPN\end{tabular}}&\textbf{709.12} & \textbf{192.71} & \textbf{1.88} & \textbf{0.82}	
\end{tabular}
\caption{Quantitative evaluation on KITTI DC Dataset. The CSPN, TWISE, DSPN, DeepLiDAR, ACMNet, CSPN++, NLSPN, GuideNet, FCFRNet, PENet mean \cite{CSPN} \cite{TWISE} \cite{9191138} \cite{8953782} \cite{9440471} \cite{CSPN++} \cite{NLSPN} \cite{guidenet} \cite{liu2021fcfr}  \cite{hu2020PENet}, respectively. }
\label{table3}
\end{table}
\subsection{Datasets}
\textbf{NYU Depth V2 dataset} \cite{NYUv2} is comprised of RGB and depth images captured by a Microsoft Kinect camera in 464 indoor scenes. Following \cite{CSPN} \cite{NLSPN} \cite{liu2021fcfr}, Our model is trained on a subset of 50K images from the official training split and tested on the 654 images from the official labeled test set. Each image is downsized to 320×240, and then 304×228 center-cropping was applied.

\textbf{KITTI DC dataset} \cite{8374553} consists of over 90K RGB and LiDAR pairs for training, 1K pairs for validation, and another 1K pairs for testing. 
The training pairs are top cropped for no LiDAR projection data (i.e., top 100 pixels), and then center cropped to 1216×256. Incidentally, the sparse depth maps are provided by HDL-64 LiDAR contain less than $6\%$ valid values, and the ground truth depth maps are generated by the accumulated LiDAR scans of multiple timestamps with about $14\%$ density. 
\subsection{Metrics}
For the NYU Depth V2 dataset, root mean square error (RMSE), mean absolute relative error (REL), and percentage of pixels satisfying $\delta _{\tau }$ are selected as the evaluation metrics.
For the KITTI DC Dataset, we adopt error metrics same as the dataset benchmark, including RMSE, mean absolute error (MAE), inverse RMSE (iRMSE), and inverse MAE (iMAE). 
All the metrics for evaluation are shown as follows.

RMSE(mm): $\sqrt{\frac{1}{v}\sum _{x}(\hat{h}_{x}-h_{x})^{2}}$

MAE(mm): $\frac{1}{v}\sum _{x}|\hat{h}_{x}-h_{x}|$

iRMSE(1/km): $\sqrt{\frac{1}{v}\sum _{x}(\frac{1}{\hat{h}_{x}}-\frac{1}{h_{x}})^{2}}$

iMAE(1/km): $\frac{1}{v}\sum _{x}|\frac{1}{\hat{h}_{x}}-\frac{1}{h_{x}}|$

REL: $\frac{1}{v}\sum _{x}|\frac{\hat{h}_{x}-h_{x}}{h_{x}}|$

$\delta _{\tau }$: $max(\frac{h_{x}}{\hat{h}_{x}}-\frac{\hat{h}_{x}}{h_{x}})< \tau ,\tau \in \{1.25,1.25^{2},1.25^{3}\}$

\begin{table*}[t]
\centering
\begin{tabular}{ccccccccccc}
\hline
\multicolumn{1}{c|}{\multirow{3}{*}{Method}} & \multicolumn{1}{c|}{\multirow{3}{*}{Iteration}} & \multicolumn{2}{c|}{DySPN Module}                                                        & \multicolumn{1}{c|}{\multirow{3}{*}{DS}} & \multicolumn{4}{c|}{Training Strategy}                                                                                                                                                         & \multicolumn{2}{c|}{Results(Lower the better)}                            \\ \cline{3-4} \cline{6-11} 
\multicolumn{1}{c|}{}                        & \multicolumn{1}{c|}{}                           & \multicolumn{1}{c|}{\multirow{2}{*}{Sigmoid}} & \multicolumn{1}{c|}{\multirow{2}{*}{Softmax}} & \multicolumn{1}{c|}{}                    & \multicolumn{3}{c|}{Weight Decay}                                                           & \multicolumn{1}{c|}{\multirow{2}{*}{\begin{tabular}[c]{@{}c@{}}Stochastic\\ Depth\end{tabular}}} & \multicolumn{1}{c|}{\multirow{2}{*}{RMSE(mm)}} & \multirow{2}{*}{MAE(mm)} \\ \cline{6-8}
\multicolumn{1}{c|}{}                        & \multicolumn{1}{c|}{}                           & \multicolumn{1}{c|}{}                         & \multicolumn{1}{c|}{}                         & \multicolumn{1}{c|}{}                    & \multicolumn{1}{c|}{$5\times 10^{5}$} & \multicolumn{1}{c|}{$10^{5}$} & \multicolumn{1}{c|}{$5\times 10^{6}$} & \multicolumn{1}{c|}{}                                                                            & \multicolumn{1}{c|}{}                          &                          \\ \hline
7×7 CSPN                                     & 12                                              &                                               &                                               &                                          &                               &                              &                               &                                                                                                  & 789.6                                          & 201.4                    \\
7×7 CSPN                                     & 6                                               &                                               &                                               &                                          &                               &                              &                               &                                                                                                  & 811.0                                          & 207.3                    \\
7×7 DySPN                                    & 12                                              & \checkmark                                    &                                               &                                          &                               &                              &                               &                                                                                                  & 768.1                                          & 196.4                    \\
-                                             & 6                                               & \checkmark                                    &                                               &                                          &                              &                              &                               &                                                                                                  & 769.0                                          & 195.9                    \\
-                                             & 6                                               &                                               & \checkmark                                    &                                          &                              &                              &                               &                                                                                                  & 770.0                                          & 197.1                    \\
-                                             & 6                                               & \checkmark                                    &                                               & \checkmark                               &                              &                              &                               &                                                                                                  & 763.9                                          & 195.5                    \\
-                                             & 6                                               & \checkmark                                    &                                               & \checkmark                               & \checkmark                   &                              &                               &                                                                                                  & 753.2                                          & 196.5                    \\
-                                             & 6                                               & \checkmark                                    &                                               & \checkmark                               &                              & \checkmark                   &                               &                                                                                                  & 752.9                                          & 194.6                    \\
-                                             & 6                                               & \checkmark                                    &                                               & \checkmark                               &                              &                              & \checkmark                    &                                                                                                  & 755.8                                          & 195.2                    \\
-                                             & 6                                               & \checkmark                                    &                                               & \checkmark                               &                              & \checkmark                   &                               & \checkmark                                                                                       & 747.0                                          & 193.2                    \\
-                                             & 6                                               & \checkmark                                    &                                               & \checkmark                               &                              &                              &                               & \checkmark                                                                                       & 745.8                                          & 192.5                    \\ \hline
\end{tabular}
\caption{Ablation study on the KITTI DC validation set. DS means diffusion suppression.}
\label{table1}
\end{table*}

\subsection{Evaluation on NYU Depth v2 Dataset}
We verified the effectiveness of our method in indoor scenes by evaluating on NYU Depth v2 Dataset. The quantitative comparison results are shown in Table. \ref{table2}. Our proposed algorithms achieve the best results and outperform other SoTA methods. Fig. \ref{fig4} visualizes the attention maps learned by our network. Qualitatively, attention maps $\pi_{1}^{t}$ increase significantly with iteration, whereas attention maps $\pi_{2}^{t}$ and $\pi_{3}^{t}$ decrease. These attention maps, such as $\pi_{0}^{3}$ and $\pi_{2}^{3}$, contain a lot of boundary information. In contrast with CSPN++ \cite{CSPN++}, we find a majority of pixels use a big kernel with a large receptive field at first several iterations for fast recovery, while in the last iteration, pixels around object and surface boundary tend to use a smaller kernel, so that the attention map $\pi_{1}^{t}$ is becoming more and more obvious. We also observe an increasing trend in the attention maps of the diffusion suppression operation(i.e. $\pi_{0}^{t}$).

\subsection{Evaluation on KITTI DC Dataset}
We also verified the effectiveness of our method in outdoor scenes by evaluating on KITTI DC Dataset. The results of quantitative comparisons are shown in Table. \ref{table3}. We finally submit the evaluation result of Deformable DySPN, which is the best of our three variations. Our DySPN outperforms all the other works with a significant improvement (nearly 21mm in RMSE) and ranks 1st on KITTI DC evaluation in all metrics (RMSE, MAE, iRMSE, and iMAE) by the time of submission. In Fig. \ref{fig5}, we compare the results of our method with several works. Our work is more consistent in some cases where the depth maps are extremely sparse.  

\subsection{Ablation Studies}
In this section, we first perform ablation experiments to verify the effectiveness of our model design. Our experiments are trained on KITTI DC annotated train dataset and tested on the selected validation dataset. A simple 7×7 CSPN \cite{CSPN} is implemented as our baseline model. Table. \ref{table1} demonstrates the effectiveness of each design proposed in our 7×7 DySPN, including the DySPN module, the diffusion suppression (DS) operation, and the network training strategies. In addition, we employ a U-net with the backbone of ResNet-34 as our base network. 

%\paragraph{The effectiveness of the dynamic spatial propagation} 
\textbf{DySPN module} and CSPN module are compared in 6 and 12 iterations respectively, as shown in Table. \ref{table1}. In particular, the RMSE is decreased from 811.0$mm$ of CSPN to 769.0$mm$ of DySPN with a significant 42$mm$ reduction when the number of iterations is 6. DySPN module also shows better convergence, because the RMSE is not affected when the number of iterations is increased from 6 to 12. We also find that there is little difference between the sigmoid layer and the softmax layer on 
our module. 
%Inspired by dynamic filter approaches such as CondConv \cite{NEURIPS2019_f2201f51} and DynamicConv \cite{9157588}, we normalize the attention maps by a sigmoid or a softmax layer to generate an adaptive affinity matrix for a DySPN module. 
%We compared the influence of the sigmoid layer and the softmax layer on DySPN module. But we find that there is little difference between the sigmoid and the softmax. 

%\paragraph{The effectiveness of the diffusion suppression operation} 
\textbf{Diffusion suppression (DS) operation} is easy to be embedded to DySPN module as proposed in Eq. \ref{eq7}. As visualized in Fig. \ref{fig4}, high thermal values gradually appear at the boundary of the attention maps $\{\pi _{0}^{1},\cdots ,\pi _{0}^{6}\}$, it clearly shows the edge-preserving effect of diffusion suppression. In Table. \ref{table1}, we observe a reduction in RMSE from 769.0$mm$ to 763.9$mm$, which means our design is successful.

%\paragraph{The effectiveness of training strategies} 
\textbf{Training strategy} of weight decay is mentioned in many depth completion works \cite{CSPN++} \cite{hu2020PENet} \cite{liu2021fcfr} \cite{guidenet} to reduce model over-fitting. In this case, we try different values of weight decay and figure out the best one as shown in Table. \ref{table1}. In addition, we adopt a stochastic depth strategy\cite{Stochastic-Depth} to play a similar role as weight decay. Experiment results demonstrate that it is better to use stochastic depth alone than with weight decay.  

\begin{table}[t]
\centering
\begin{tabular}{ccccc}
\hline
SPN Modules      & Iter. & \begin{tabular}[c]{@{}c@{}}Time\\ (ms)\end{tabular} & \begin{tabular}[c]{@{}c@{}}RMSE\\ (mm)\end{tabular} & \begin{tabular}[c]{@{}c@{}}MAE\\ (mm)\end{tabular} \\ \hline
NLSPN            & 18    & 55                                               & -                                                   & -                                                  \\
PENet(C2)        & 12    & 15                                               & 757.2                                               & 209.0                                              \\
7×7 CSPN        & 12    & 70                                               & 789.6                                               & 201.4                                              \\ \hline
7×7 DySPN       & 6     & 38                                               & 745.8                                               & 192.5                                              \\
Dilated -      & 6     & 6.9                                              & 748.5                                               & 193.8                                              \\
Deformable -       & 6     & 10.3                                               & 739.4                                               & 191.4                                              \\ \hline
\end{tabular}
\caption{Runtime and preformance of three variants of DySPN on the KITTI DC validation set. }
\label{table4}
\end{table}

%\paragraph{The effectiveness of Dilated DySPN and Deformable DySPN} 
%Inspired by PENet \cite{hu2020PENet}, DSPN \cite{9191138} and NLSPN \cite{NLSPN}, we propose Dilated DySPN and Deformable DySPN. 
\textbf{Dilated DySPN and Deformable DySPN} improve the effect and speed with fewer iterations and fewer neighbor samples, as illustrated in Table. \ref{table4}. 
Specifically, Dilated DySPN module is designed for a better speed-accuracy trade-off. It shows not only over $2\times$ faster than the C2 module of PENet but also improves RMSE from 757.2$mm$ to 748.5$mm$. Deformable DySPN is improved on precision. As summarized in Table. \ref{table3}, it ranks 1st on the KITTI DC benchmark at the time of submission.

\section{Conclusion}
In this paper, we proposed DySPN with a non-linear propagation model, which applies spatial and sequential attention to generate a series of adaptive affinity matrices. Furthermore, we presented a diffusion suppression operation to avoid the over-smoothing problem. To reduce complexity, we elaborated on three variants, including 7×7 DySPN, Dilated DySPN, and Deformable DySPN, which have different speed-accuracy trade-offs. Our ablation studies verified the superior performance of our DySPN variations on both indoor and outdoor depth completion datasets. By the way, we believe that DySPN can not only be used for image-guided depth completion but also be applied to other aspects, such as image segmentation and cellular automata modeling.

%\section{Acknowledgments}

% Use \bibliography{yourbibfile} instead or the References section will not appear in your paper
%\nobibliography{aaai22}
\bibliography{aaai22.bib}

\bigskip
%\noindent Thank you for reading these instructions carefully. We look forward to receiving your electronic files!

\end{document}